\documentclass[letterpaper, 10 pt, conference]{ieeeconf}

\title{
  \LARGE \bf

  Synchronized Dual-arm Rearrangement via Cooperative mTSP  
  }

  \author{Wenhao Li$^{1}$, Shishun Zhang$^{1}$, Sisi Dai$^{1}$, Hui Huang$^{2}$, Ruizhen Hu$^{2,*}$, Xiaohong Chen$^{3,4,*}$, Kai Xu$^{1,*}$ \\
$^{1}$National University of Defense Technology  $\quad$
$^{2}$Shenzhen University$\quad$ 
$^{3}$Xiangjiang Laboratory$\quad$\\
$^{4}$Hunan University of Technology and Business$\quad$
$^{*}$Corresponding Authors}

\usepackage{graphicx}
\usepackage{amsmath}
\usepackage{amssymb}
\usepackage{multirow}
\usepackage{xcolor}
\usepackage{wrapfig}
\usepackage{caption}
\usepackage{diagbox}
\begin{document}

\maketitle







\begin{abstract}
Synchronized dual-arm rearrangement is widely studied as a common scenario in industrial applications. It often faces scalability challenges due to the computational complexity of robotic arm rearrangement and the high-dimensional nature of dual-arm planning. To address these challenges, we formulated the problem as cooperative mTSP, a variant of mTSP where agents share cooperative costs, and utilized reinforcement learning for its solution. 
Our approach involved representing rearrangement tasks using a task state graph that captured spatial relationships and a cooperative cost matrix that provided details about action costs.
Taking these representations as observations, we designed an attention-based network to effectively combine them and provide rational task scheduling.
Furthermore, a cost predictor is also introduced to directly evaluate actions during both training and planning, significantly expediting the planning process. Our experimental results demonstrate that our approach outperforms existing methods in terms of both performance and planning efficiency.

\end{abstract}


\section{Introduction}
The rising need for flexible and adaptable systems in the industrial sector has amplified the research importance of synchronized dual-arm rearrangement, a scenario that is frequently encountered in areas like sorting, transportation, and manufacturing \cite{p2,p3}. Coordinating two arms within a shared workspace not only boosts work efficiency through parallelism but also enables a more flexible completion of rearrangement tasks \cite{p10}. Additionally, with a reduced state space, synchronous planning can effectively avoid deadlock issues compared with asynchronous planning, resulting in higher stability \cite{p7}.

However, dual-arm rearrangement poses computational complexity primarily due to its integration of discrete and continuous reasoning. It involves both an upper-level task scheduling problem and a lower-level trajectory planning problem that are intricately connected to each other \cite{rearrangement}. 
In the upper-level scheduling, the objective is to identify the optimal action sequence to minimize global execution costs. However, during the scheduling process, lower-level continuous trajectory planning is necessary for each discrete action to ensure collision avoidance and assess its actual execution cost, which demands a substantial amount of computational resources.
Moreover, coordinating two arms in a shared workspace, in contrast to single-arm, results in an exponential expansion of the discrete task action space and a doubling of the dimensionality of robot arm configurations \cite{p8, drrt}. This leads to a considerably larger search space and a more complex evaluation process for task scheduling, which contributes to a scalability challenge for dual-arm rearrangement when faced with a large number of tasks \cite{p4}. 

Current multi-robot task and motion planning (MRTAMP) methods employ various strategies to enhance the efficiency of task scheduling \cite{p2,p3,p4,p5,p9,p10,p11,p12,p13,p14}. Nevertheless, as long as it requires frequent communication between task allocator and motion planner, scalability issues remain a persistent challenge.
With regard to this challenge, our perspective is that, concerning task scheduling, its interaction with trajectory planning is solely aimed at obtaining evaluations for all potential actions, while it does not specify any evaluation process. If we can directly obtain the accurate costs of actions, solving the dual-arm rearrangement problem in a synchronous setting can effectively be transformed into addressing a variant of the multiple Traveling Salesmen Problem (mTSP) \cite{mtsp-survey}, where agents share common cooperative costs. Given that learning-based methods have already shown strong performance in solving multi-agent scheduling problems \cite{p15,p16,p17,p18}, leveraging attention-based reinforcement learning to address this variant problem not only ensures scalability for a large number of tasks but also offers a level of flexibility and adaptability to the policy through online planning \cite{p19}.

In this paper, we formalize the synchronized dual-arm rearrangement problem as cooperative mTSP, a variant of mTSP where agents share cooperative costs, and utilize attention-based reinforcement learning to overcome the scalability challenge. To achieve this, we develop a network that can compile the cooperative costs associated with mTSP and provide rational scheduling. Furthermore, we introduce a cost predictor capable of directly estimating cooperative costs for all potential actions based on current state information. This decouples the continuous geometric reasoning aspect of the rearrangement problem
and accelerates both the training and planning processes. Our experiments demonstrate that our approach offers several advantages over existing methods. Even with training on a limited number of tasks, the model can effectively generalize to a larger number and provide high-quality task scheduling in a very short time.

\begin{figure*}[t]
\centering{\includegraphics[width=0.95\textwidth]{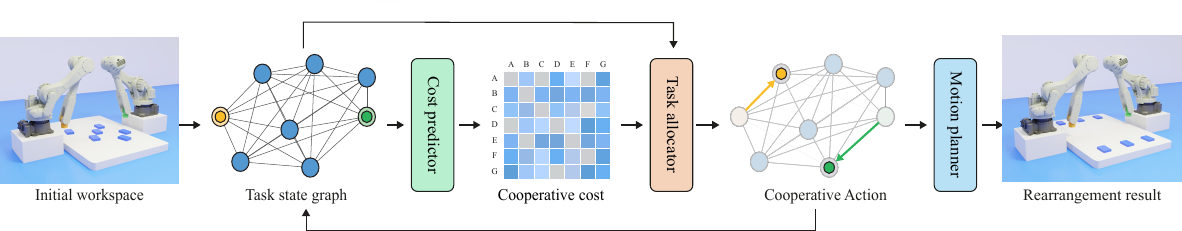}}
\caption{Our algorithm comprises an RL-based task allocator and a cost predictor. When given a rearrangement task, we translate it into a task state graph and employ the cost predictor to calculate the cooperative cost matrix. Both the state graph and cost matrix are provided to the task allocator to determine current action. The action generated at each step is used to update the task state graph for the next iteration, continuing until the rearrangement task is completed and the final action sequence is obtained.}
\label{pipeline}
\end{figure*}


\section{Related Works}
\subsection{Multi-robot task allocation for rearrangement problem}
Multi-robot Task and Motion Planning (MRTAMP) provides a framework for combining discrete and geometric reasoning, while it faces scalability and solution quality challenges \cite{p4}. Existing methods to address scalability in such tasks can be categorized into two classes:
The first class aims to reduce the search space, thereby decreasing the number of calls to the motion planner and saving time. 
For example, Gao et al. \cite{p5} use a heuristic method to obtain high-quality initial solutions, which are then adjusted based on actual collision problems. Pan et al. \cite{p4} introduce a scheduler layer between the task planner and motion planner to prune the search space and reduce the calls to the motion planner. Hartmann et al. \cite{p24} reduce the search space by priority-based planning. Shome et al. \cite{p2} implement a perfect matching with the minimum transfer cost and then optimize the order based on the matching. However, these methods still struggle when dealing with a substantial number of tasks because the search space for dual-arm planning rapidly expands as number of tasks increases.
The second class aims to simplify the action evaluation process during planning by directly using of pre-computed results (such as roadmaps, cost, or collision information) during planning to improve efficiency \cite{p3,p11,p12,p14}. However, recalculations are needed when tasks or environments change, which is time-consuming and challenging to adapt to other scenarios. Additionally, some methods use different representations to calculate costs instead of trajectory planning. Yoon et al. \cite{p9} use the shadow space, and Behrens et al. \cite{p13} use voxels swept by the arms to replace trajectory planning and calculate the overlap area to represent costs. Gafur et al. \cite{p10} treat the arm as line segments or spindles and consider no intersection between segments as a hard constraint for motion planning. However, overly simplified action evaluation can lead to a significant decrease in overall planning performance, and the design of different representations also incurs design costs. 

There are also some learning-based methods focused on supporting rearrangement tasks, but mainly for single-arm operations \cite{p1,p21}. Furthermore, some works use learning-based methods to solve dual-arm collaborative tasks while they do not prioritize task scheduling \cite{p20,p26,p32,p22}.

Our approach addresses dual-arm collaborative rearrangement tasks by using attention-based reinforcement learning for task assignment, which ensures that planning time increases linearly with the number of tasks. Additionally, a neural-based predictor accurately predicts task action costs, allowing us to bypass the trajectory planning, greatly reducing the time required while minimizing performance loss.

\subsection{Reinforcement learning for solving mTSP}
In recent years, learning-based approaches have gained a lot of attention in mTSP. Compared to non-learning methods, using attention-based reinforcement learning (RL) policies can solve large-scale mTSP problems in a very short time, making it applicable to many real-time tasks\cite{3dbpp,3dbpp0,ibs}. Most neural-based methods approach the mTSP in a decentralized manner. For instance, Park et al. \cite{p16} use a decentralized network based on graph attention to achieve end-to-end solutions for both mTSP and mJSP. Cao et al. \cite{p15} propose a fully parameter-sharing decentralized approach that generalized policies to a large number of agents. Gao et al. \cite{p18} improve learning efficiency by introducing gated transformers. Zhang et al. \cite{p33} use multi-agent reinforcement learning to solve the multivehicle routing problem (mVRP) with a time window, which is a variation of mTSP, where each agent has capacity and time constraints.

In our variant of the mTSP, however, we have only two agents, but they share cooperative costs and feasibility. We are not primarily concerned with generalizing to a larger number of agents. Instead, we focus on enhancing the collaborative decision-making capabilities between these two agents, which existing methods do not specifically address.

\section{Method}

\begin{figure*}[t]
\centering{\includegraphics[width=0.95\textwidth]{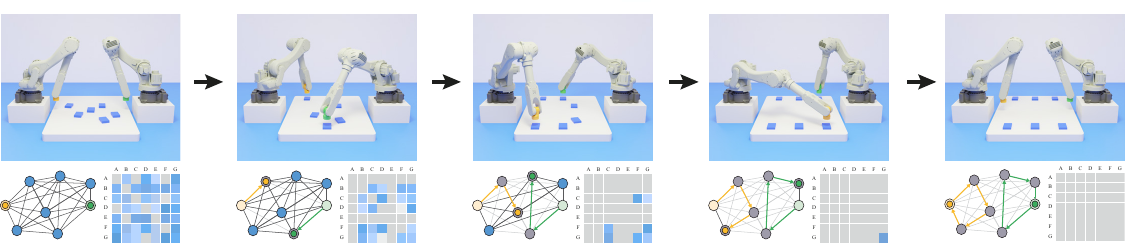}}
\caption{We formalize the rearrangement task into the cooperative mTSP with a task state graph and a cooperative cost matrix dynamically updated with interactions. The state graph contains spatial information between tasks and agents, while the cost matrix represents the cost of all potential joint actions for two agents at the current state. Each time dual arms execute a joint action, the state graph is updated accordingly, and the cost matrix is recalculated.}
\label{formulation}
    
\end{figure*}

    

Fig. \ref{pipeline} shows an overview of our framework for solving dual-arm rearrangement problems in a synchronous setting. Given $n$ objects resting on a tabletop workbench, our goal is to find the optimal sequence of actions for two robotic arms to rearrange them to their target positions with the shortest execution time. To achieve this, we formalize the problem as cooperative mTSP, a variant of mTSP where agents share cooperative costs, represented by a task state graph and a cooperative cost matrix, and employ a reinforcement learning (RL) policy to provide reasonable task assignments. 

Our algorithm consists of a task allocator based on an attention-based RL policy and a cost predictor based on a multi-layer perceptron (MLP). Presented with a rearrangement task, we begin by converting the environmental state into a task state graph designed for the mTSP. With this graph, a cooperative cost matrix is calculated by directly estimating cooperative costs for all potential joint actions of two arms at the current step using the cost predictor. Following it, both the task state graph and the predicted cost matrix are fed into the task allocator to infer a suitable arrangement action with the aim of minimizing the total cost. The selected action will be used to update the task state graph for the next iteration. This iterative process continues until both agents successfully complete the rearrangement task, providing us with a complete sequence of actions. For simplification, we consider that all objects are of uniform box size and that their initial and target positions do not overlap.

\subsection{Formulation}
We transform the dual-arm rearrangement scenario into a MinMax multi-depot multiple traveling salesman problem with cooperative costs, represented by a task state graph and a dynamic cooperative cost matrix dynamically updated with interactions. The task state graph captures spatial relationships between tasks and agents, while the cooperative cost matrix details costs for all potential actions in the current workspace state.

\textbf{Task state graph.} 
The state graph is constructed as a non-fully connected graph that reflects the spatial relationship and task completion status within the current workspace. It consists of $n$ task nodes and $2$ depot nodes with agents positioned at them. Two depot nodes are not connected to each other since agents will not visit the opponent’s depots. 
The task nodes
represent the 'cities to be visited' in mTSP, with each node corresponding to an arrangement task for a single object. Each task node contains information about both the starting pick pose and ending placement pose for object arrangement. These details are represented in both absolute Cartesian coordinates for position and Euler angles for orientation as $s_{i} = [p^{s}_{i}, o^{s}_{i}, p^{e}_{i}, o^{e}_{i}] (i \in {1,\dots,n})$, where $i$ represents the index of arrangement task. 
On the other hand, the depot nodes 
serve as starting points for two agents. They share a comparable representation structure with task nodes, containing information about the initial and current poses of the robotic arms' end-effectors as $s_{k} = [p^{i}_{k}, o^{i}_{k}, p^{c}_{k}, o^{c}_{k}] (k \in {r_1,r_2})$, where $r_{1},r_{2}$ stand for two robotic arms. 

\textbf{Cooperative cost matrix.} 
The cost matrix represents cooperative costs for all potential actions in the current workspace state. 
These costs are influenced by both task-related information and the current configuration of agents during action execution. Therefore, as agents perform rearrangement tasks, the cooperative cost matrix dynamically changes with the current workspace state as well. The cost matrix consists of $(n+1)*(n+1)$ grid cells, which are aligned with the joint action space of two agents. Each grid cell contains two types of cost associated with relevant action, denoted as 
$c^{mv}$ and $c^{tf}$.
They represent the cost associated with coordinated movements for moving to the objects
and transferring them to their target poses
, which are calculated from the configuration of tasks and arms at the current step through generating and evaluating relative motion trajectories. In cases of unfeasible actions, like those allocating the same tasks or completed tasks, costs will be set to infinity.

\textbf{Cooperative mTSP.}
Fig. \ref{formulation} provides an overview of the entire process of formalizing synchronized dual-arm rearrangement as the problem of solving the cooperative mTSP. In this problem, agents are tasked with discovering a feasible sequence of action pairs to visit all task nodes and return to their respective depot nodes while minimizing the total time. 

In each decision-making step, agents must select an arrangement task for synchronized execution based on the current state graph and cost matrix. This selection can be represented as the indices of a pair of task nodes in the state graph, denoted as $[I^{1}_{t}, I^{2}_{t}]$. Furthermore, cost of the executed action can be calculated using the current state's cooperative cost matrix as $c^{mv}_{t}(I^{1}_{t}, I^{2}_{t}) + c^{tf}_{t}(I^{1}_{t}, I^{2}_{t})$, which represents the time required for two arms to perform synchronized moving and transfer movements for task $I^{1}_{t}$ and $ I^{2}_{t}$ at time step $t$. The time required for picking and placing objects is disregarded as it can be considered as a fixed constant. 

Following the execution of an arrangement task, agents within the task state graph are expected to move to the relevant task nodes, marking the nodes that have been completed. Subsequently, the cooperative cost matrix is recalculated based on the information from the updated task state graph, reflecting changes in the dual-arm configuration. This iterative process continues until all arrangement tasks are completed. Finally, both agents return to their depots to conclude their tours, with the time cost denoted as $c^{rt}$, which dynamically changes based on the location of the last completed task.
Our primary objective is to determine a feasible cooperative action sequence for both agents to complete their path pairs while minimizing the cumulative cost, quantified as follows:
\begin{equation}
C = \sum^{n/2}_{t = 1} c^{mv}_{t}(I^{1}_{t}, I^{2}_{t}) +  \sum^{n/2}_{t = 1} c^{tf}_{t}(I^{1}_{t}, I^{2}_{t}) + c^{rt}
\end{equation}

\begin{figure*}[t]
\centering{\includegraphics[width=0.98\textwidth]{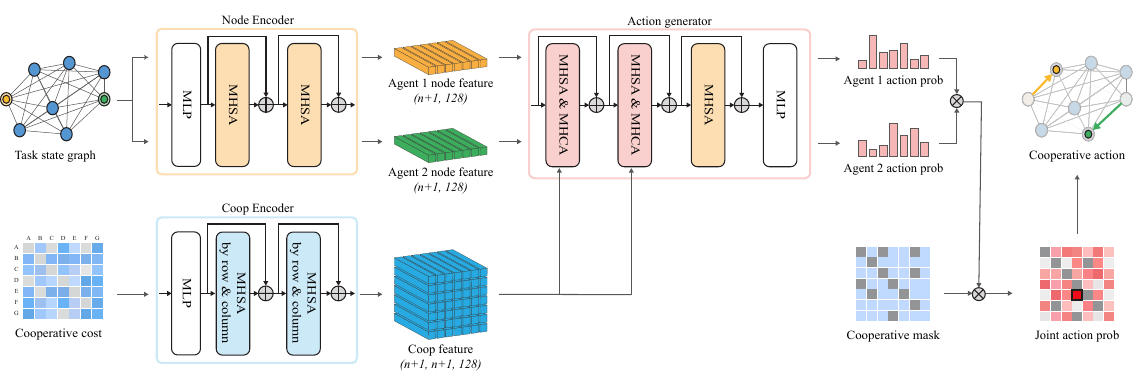}}
\caption{Our network consists of a node encoder, a coop encoder, and an action generator. The node encoder processes the state graph to encode spatial information, while the coop encoder takes the cost matrix as input to encode cooperative information. The action generator combines both encoded features and generates a probability map for all potential actions.}
\label{net-archn}
\end{figure*}


\subsection{Task allocator}
Our task allocator consists of a decentralized attention-based RL policy, where two agents have shared network parameters. Its network architecture is shown in Fig. \ref{net-archn}.

\textbf{Observation.}
We provide both agents with the task state graph and cooperative cost matrix as observations, ensuring a fully observable world. For the task state graph, we incorporate relative position and orientation with respect to each agent's end-effector and concatenate them with the absolute features to enhance policy robustness. Concerning the cooperative cost matrix, we augment it with additional masks to describe the feasibility of all potential actions. Identical cost matrices are provided to both agents with rows and columns transposed for alignment purposes.

\textbf{Action.}
Our policy generates action sequences iteratively until either all tasks are completed or an unsolvable situation arises. At each time step, it generates a probability map for all potential joint actions. This map is calculated by multiplying the individual action probabilities from both agents and is filtered through a global mask. The action with the highest probability is selected as the current output.

\textbf{Reward design.}
In terms of reward design, we adopted a dense reward with a sparse failure penalty to incentivize policy decisions. For each time step $t$, the reward function is designed as follows:
\begin{equation}
    R_{t} = - {(c^{mv}_{t}+c^{tf}_{t})/2 - c_{p}}
\end{equation}


Here, $(c^{mv}_{t}+c^{tf}_{t})/2$ represents the dense reward, signifying the action cost at time step $t$. Additionally, we use an adaptive failure penalty, denoted as $c_{p}$, as a sparse reward whose value varies depending on the number of undone tasks at episode termination. When the episode is not terminated, $c_{p}$ is set to zero. However, if the episode is terminated, $c_{p}$ is adjusted to $n_{u}/n$, with $n_{u}$ representing the number of undone tasks at termination.
With this reward setup, our policy, combined with step-back techniques, strives to optimize cumulative costs whenever feasible, while also taking precautions to avoid unsolvable situations.

\textbf{Network Architecture.}
As illustrated in Fig. \ref{net-archn}, our network comprises three primary components: node encoder, coop encoder, and action generator. Agents process observation from the task state graph to gather spatial relationships between tasks and agents via the node encoder. Concurrently, they collect cooperative relationship information between agents from the cooperative cost matrix using the coop encoder. These two sets of compiled features are then integrated within the action generator, resulting in individual action probabilities for each agent.

The node encoder consists of an MLP layer and two multi-head self-attention (MHSA) layers. Taking observation from the task state graph as input, we initially employ an MLP layer to embed the input into a 128-dimensional feature vector. All node features then pass through MHSA layers for information aggregation. Residual connections and layer normalization are applied between attention layers to ensure a better convergence. Notably, since agents do not reach the depot of the opposing agent, the node feature representing the depot node of the opposing agent is excluded from the output. This results in n+1 node features, which are then fed into the action generator to generate action probabilities.

The coop encoder has the same structure as the node decoder. However, it employs a distinct attention mechanism within the attention layers, which we refer to as "MHSA by row and column". In the joint action space of agents, not all actions exhibit strong correlations. Aggregating information from all actions into a single action would result in information redundancy. Therefore, in MHSA by row and column, we aggregate feature information specifically along the rows and columns of each matrix cell. This choice is made because these rows and columns represent the strong correlations associated with the joint actions where one agent's action aligns with the matrix cell.

The action generator integrates the encoded node and coop features and outputs action probabilities for each agent. It employs two layers of multi-head self and cross-attention (MHSA\&MHCA) to fuse these two types of information. 
In the self-attention component, node feature information is aggregated in the same approach as the node encoder. However, in cross-attention, to avoid redundancy, each node aggregates only the row of coop features relevant to itself rather than all features in encoded coop features.
After that, integrated features are processed through a masked self-attention layer and an MLP layer, followed by a Softmax operation to output the action probabilities for the agent.

\subsection{Cost predictor}
To expedite the training of the policy and planning process, we utilized a pre-trained model as a cost predictor to directly estimate cooperative costs for all potential actions at each step. Our cost predictor consists of a three-layer 512-dimensional MLP. With the current and target configurations of the dual robotic arms as well as the size of the grasping object's bounding box as input, it provides predictions regarding the feasibility of “transit” and “transfer” actions and estimates the execution time of action trajectories. These predictions are used to construct the mask and cost components of the cooperative cost matrix, which serves as coop observation input to the task allocator.
In our specific setup, we trained the predictor using one million randomly generated “transit” and “transfer” tasks in a simulated environment. The predictor achieves an accuracy rate of over 98\% in mask prediction, with prediction errors for costs being less than 0.5\%.

\subsection{Motion planner and action execution}
Our algorithm doesn't impose any other specific requirements on the motion planner as long as we can learn a cost predictor that can accurately predict the cost of actions executed by the motion planner. 
 When the motion planner fails to generate a valid path or the task allocator leads to an unsolvable situation, we employ the step-back approach. This allows us to revert to the task state graph prior to the failure, mask out the actions that caused the failure, and re-plan the actions, which significantly increases the success rate of planning. The online planning capability of reinforcement learning makes this entire process feasible and efficient.

\section{Experiment and Result}
\subsection{Experiment setting}
We employ the Proximal Policy Optimization (PPO) \cite{ppo} algorithm and train our RL policy using randomly generated task data. In our experiment, all algorithms are trained and evaluated on a single NVIDIA 3080TI GPU paired with an Intel i-10980XE CPU. We utilize the GPU for training and evaluating all learning-based methods, as well as training and utilizing the cost predictor. For evaluating other baselines, we employ a single-core CPU.

Our experiments are conducted in a PyBullet simulation environment. In this setting, multiple box-shaped objects are positioned on a flat working surface, with their initial and target positions non-overlapped. Two KAWASAKI RS050N robotic arms are situated on opposite sides, executing synchronized rearrangement tasks from their respective starting positions. For safety considerations, the dual arms are prohibited from simultaneously picking or placing two objects within a distance of less than 0.1 meters. To facilitate comparisons, we disregard the cost associated with fixed-duration grasping and placement processes, focusing solely on calculating the total time spent during all transit and transfer processes, which we refer to as the cumulative cost.

We create several task datasets by randomly sampling based on the number of parts, with each dataset containing one hundred task data for testing. We assess algorithm performance using three metrics: single-shot success rate, average cumulative cost, and average planning time.

\begin{table*}[th]
\centering
\caption{Performance comparison between our method and baselines. Single-shot success rate, average cumulative cost, and average planning time are used as metrics}
\resizebox{\textwidth}{!}{
\renewcommand{\arraystretch}{1}

\begin{tabular}{l||ccc|ccc|ccc|ccc|ccc}
\hline
method                 & \multicolumn{3}{c|}{n = 6}  & \multicolumn{3}{c|}{n = 10}       & \multicolumn{3}{c|}{n = 20}       & \multicolumn{3}{c|}{n = 30}       & \multicolumn{3}{c}{n = 40}         \\ \hline
metrics                & succ & cost  & time         & succ & cost  & time               & succ & cost  & time               & succ & cost  & time               & succ & cost   & time               \\ \hline
Exhaust Search(GT)     & 0.99 & 18.15 & 68.72        & 1    & 25.81 & \textgreater{}1000 & \scalebox{0.85}[1]{$\times$}    & \scalebox{0.85}[1]{$\times$}     & \textgreater{}1000 & \scalebox{0.85}[1]{$\times$}    & \scalebox{0.85}[1]{$\times$}     & \textgreater{}1000 & \scalebox{0.85}[1]{$\times$}    & \scalebox{0.85}[1]{$\times$}      & \textgreater{}1000 \\
Perfect Matching       & 0.91 & 19.60 & 17.70        & 0.89 & 28.87 & 49.20              & 0.84 & 50.82 & 196.79             & 0.86 & 72.30 & 442.75             & 0.91 & 92.70  & 787.80             \\
Task Allocator(mTSP)   & 0.69 & 20.85 & \textless{}1 & 0.67 & 31.61 & \textless{}1       & 0.7  & 58.70 & \textless{}1       & 0.66 & 85.49 & 2.40               & 0.7  & 110.12 & 4.49               \\
Task Allocator(Greedy) & 0.99 & 20.15 & 3.68         & 1    & 29.82 & 5.91               & 1    & 51.95 & 12.07              & 1    & 72.97 & 16.33              & 1    & 99.42  & 29.16              \\
Task Allocator(Ours)   & 0.99 & \pmb{19.23} & 4.26         & 1    & \pmb{27.29} & 6.27               & 1    & \pmb{49.24} & 11.42              & 1    & \pmb{70.10} & 17.14              & 1    & \pmb{90.25}  & 23.10              \\ \hline
\end{tabular}}

\label{result}
\end{table*}

\subsection{Baseline algorithms}
We adopt the following methods as baselines:
\begin{itemize}
    \item Exhaustive Search (Ground Truth): This method involves an exhaustive search of all possible action sequences to find the optimal solution.
    \item Perfect Matching Heuristic \cite{p2}: This method selects pairs of parts from all components with the smallest transfer cost and then optimizes the sequence of pairs using classic TSP methods. Compared to traditional MILP methods, it can quickly find high-quality solutions, performing as the state-of-the-art method for dual-arm rearrangement in synchronous monotonous settings. We employ Gurobi \cite{gurobi} to identify the optimal pairs and utilize OR-tools \cite{ortools} to optimize the sequence.
    \item Task Allocator (mTSP): To highlight the substantial impact of cooperation constraints among agents in mTSP, we directly apply algorithms designed for classical mTSP to make comparisons. We choose ScheduleNET \cite{p16}, a GAT-based mTSP solver, trained with predicted cumulative cost and failure penalty as rewards.
    \item Task Allocator (greedy): To evaluate the performance of the RL policy, we replace our RL policy with a greedy policy, serving as a new baseline. The greedy policy selects the action with the lowest cooperative cost within the cost matrix as output. In the event of unsolvable situations, it resorts to step-back and replanning.
\end{itemize}

We employ a consistent motion planner across all algorithms and incorporate a global mask to quickly filter out infeasible actions during planning. This mask is determined by assessing whether both arms in the initial pose and goal pose collide in the simulation for each action. While it may not perfectly reflect feasibility during action execution using the motion planner, it offers computational efficiency as it doesn't involve generating actual trajectories.

\subsection{Experiment result}

We trained our policy in the n=10 dataset and applied it to all cases to compare with other baselines. The result is shown in Table \ref{result}. As we can see, the exhaustive search method ensures optimality while becoming impractical for more than 10 tasks due to its high computational complexity. The Perfect matching method exhibits good performance on a vast number of tasks, but its computation time also increases noticeably as the number of parts grows. Its lower success rate is attributed to situations where tasks encounter unsolvable scenarios during sequential optimization based on a pre-determined matching.
The classic mTSP method performs poorly on both success rate and cumulative cost in all datasets, as it does not account for the cooperative constraints between agents in the solution. Our approach, along with the greedy method, shows that using a predictor with the step-back technique effectively improves planning success rates. Our method achieves the best performance in terms of cumulative cost compared to other baselines, except for exhaustive search. 
Importantly, our algorithm demonstrates strong generalization performance. Even when trained on a dataset with a small number of tasks, it remains competitive when applied to a larger dataset with only a linear increase in planning time.

\begin{table}[b]
\caption{Performance comparison for ablation study. Average cumulative cost is used for evaluation}
\resizebox{\columnwidth}{!}{ 
\renewcommand{\arraystretch}{1}

\begin{tabular}{l||ccc}
\hline
method                                 & n = 10 & n = 20 & n = 40 \\ \hline
Coop encoder (MLP)                     & 28.60  & 51.96  & 96.39  \\
Coop encoder (MHSA)                    & 28.16  & 51.42  & 98.00  \\ \hline
Action generator (MLP )                & 29.89  & 53.35  & 100.12 \\
Action generator (MHSA)                & 29.15  & 52.63  & 98.28  \\
Action generator (MHCA)                & 29.15  & 51.89  & 95.84  \\ \hline
Reward with fixed penalty ($c_{p} = 1$)       & 30.18  & 52.32  & 95.98  \\
Reward with no penalty ($c_{p} = 0$)       & 30.63  & 55.03  & 102.10 \\ \hline
Euclidean distance                     & 32.45  & 60.58  & 112.04 \\
Euclidean distance + Overlap space     & 29.86  & 54.59  & 99.07  \\
Cost predictor (2-layer, 64-dimension) & 28.58  & 52.05  & 96.37  \\ \hline
Ours                                   & \pmb{27.29}  & \pmb{49.24}  & \pmb{90.25}  \\ \hline
\end{tabular}

}
\label{ablation}
\end{table}

\subsection{Ablation study}
In this paper, we conducted ablation experiments from three aspects to demonstrate the effectiveness of our design, and the results are shown in Table \ref{ablation},

\textbf{Network architecture.}
To evaluate the efficiency of our network design, we performed ablation experiments on both the coop encoder and action generator components. Regarding the coop encoder, we employed an MLP encoder and a classic MHSA encoder, each with the same number of layers as the baselines. For the action generator, we compared it to baselines replacing MHSA \& MHCA with MLP, MHSA, and MHCA.
As we can see, both of our designs are effective in aggregating information and reducing redundancy, resulting in better performance.

\textbf{Reward design.}
In our ablation on reward design, we primarily focused on the impact of sparse failure penalties, since we empirically found that modifications to dense rewards had minimal effects on performance. We conducted experiments comparing two policies: one with a fixed failure penalty ($c_{p} = 1$) and the other without any failure penalty. The results demonstrate that varying penalty settings significantly affect final performance.
A fixed large penalty tends to make the policy prioritize avoiding failures compared with optimizing the path, while no penalty can lead the policy to reduce accumulation costs by failing early. Our adaptive design, combined with step-back, better balances the objectives of finding feasible solutions and optimizing paths.

\textbf{Cost Predictor.}
To demonstrate the impact of an accurate cost predictor on policy performance, we prepared several baselines for comparison. The first baseline uses Euclidean distance directly as the execution cost. The second one employs a manually designed cost metric, which calculates the shadow spaces \cite{p9} swept by two arms and considers the overlapping area in addition to Euclidean distance as the cost. A third baseline utilizes a smaller prediction model (two-layer 64-dimensional MLP), which has an accuracy of 94.8\% in feasibility prediction and an error of 2.5\% in cost prediction. 
The results demonstrate that an accurate prediction model substantially improves policy performance across all datasets. This is mainly because our task allocator is trained using predicted costs as rewards. The closer the predicted cost aligns with the actual cost, the smaller the transfer error when deploying policy into real-world environment.

\section{Conclusion}
In this paper, we tackled the scalability challenge in synchronized dual-arm rearrangement by formulating the problem as a cooperative mTSP. We introduced an attention-based RL policy to solve the problem and a cost predictor to accelerate training and planning.
Our approach offers high-quality solutions with linear planning time, with the ability to generalize from a small number of tasks to larger ones, making it well-suited for real-time planning in diverse scenarios such as industrial rearrangement tasks which often involve standardized arm positions and grasping poses.
While our method is applicable to monotone settings, our future work will focus on employing a precedence graph and a temporal buffer to tackle non-monotone problems \cite{tap, dependency_graph}.


\section*{Acknowledgment}
 This work was supported in part by the NSFC (62325211, 62132021, 62322207), the National Key Research and Development Program of China (2018AAA0102200), the Major Program of Xiangjiang Laboratory (23XJ01009), Guangdong Natural Science Foundation (2021B1515020085), Shenzhen Science and Technology Program (RCYX20210609103121030).


\newpage
\bibliographystyle{IEEEtran}
\small\bibliography{reference}

\end{document}